\documentclass{article}

\PassOptionsToPackage{square,numbers}{natbib}

\usepackage{arxiv}

\RequirePackage{natbib}

\usepackage[utf8]{inputenc} 
\usepackage[T1]{fontenc}    
\usepackage{hyperref}       
\usepackage{url}            
\usepackage{booktabs}       
\usepackage{amsfonts}       
\usepackage{nicefrac}       
\usepackage{microtype}      
\usepackage{lipsum}
\usepackage{graphicx}
\graphicspath{ {./images/} }

\usepackage{xcolor}         
\usepackage{amsmath}

\title{Transfer learning on multi-dimensional data: a novel approach to neural network-based surrogate modeling}

\author{
Adrienne M. Propp \\
  Institute for Computational and Mathematical Engineering\\
  Stanford University\\
  Stanford, CA 94305 \\
  \texttt{propp@stanford.edu} \\
   \And
 Daniel M. Tartakovsky \\
  Department of Energy Science and Engineering\\
  Stanford University\\
  Stanford, CA 94305 \\
  \texttt{tartakovsky@stanford.edu}
}

\begin{document}
\maketitle
\begin{abstract}
The development of efficient surrogates for partial differential equations (PDEs) is a critical step towards scalable modeling of complex, multiscale systems-of-systems. Convolutional neural networks (CNNs) have gained popularity as the basis for such surrogate models due to their success in capturing high-dimensional input-output mappings and the negligible cost of a forward pass. However, the high cost of generating training data --- typically via classical numerical solvers --- raises the question of whether these models are worth pursuing over more straightforward alternatives with well-established theoretical foundations, such as Monte Carlo methods. To reduce the cost of data generation, we propose training a CNN surrogate model on a mixture of numerical solutions to both the $d$-dimensional problem and its ($d-1$)-dimensional approximation, taking advantage of the efficiency savings guaranteed by the curse of dimensionality. We demonstrate our approach on a multiphase flow test problem, using transfer learning to train a dense fully-convolutional encoder-decoder CNN on the two classes of data. Numerical results from a sample uncertainty quantification task demonstrate that our surrogate model outperforms Monte Carlo with several times the data generation budget.
\end{abstract}


\section{Introduction}
Many problems of great scientific interest are best described as systems-of-systems,  each subsystem operating at different spatial and temporal scales and governed by different physical laws. The partial differential equations (PDEs) that describe any of these subsystems may incur high computational costs when solved with conventional numerical methods, rendering uncertainty quantification (UQ) for the larger system prohibitively expensive. A salient example is the climate system, which is comprised of a number of complex subsystems and characterized by high levels of uncertainty, but for which a means of data-driven decision making is urgently needed. The prohibitively high computational cost of classical numerical schemes in such contexts has ignited an effort to develop efficient techniques for solving PDEs and estimating the expected distributions of quantities of interest. One promising approach is surrogate modeling, whereby a low-cost approximation capturing the statistical behavior of the system serves as a substitute for the full PDE model in UQ tasks \citep{kudela_recent_2022}. This approach has garnered increasing attention in recent years as a potential solution to the computational challenges posed by complex scientific systems.

A number of strategies for surrogate modeling have been proposed --- for example, Kriging and Gaussian processes \cite{kriging,krige1951statistical,rasmussen2006gaussian}; polynomial chaos expansion \cite{pce,chaos_soize}; support vector regression \cite{shi_multi-fidelity_2020}; and, recently, deep neural networks \cite[e.g.,][]{tripathy2018deep}. Surrogates based on neural networks (NNs), and convolutional neural networks (CNNs) in particular, are a natural choice for this task due to their ability to handle high-dimensional input-output mappings \cite{raghu2017expressive} and the negligible cost of a single forward pass. CNNs have proven particularly successful in image processing tasks \cite{chai_deep_2021,guo_deep_2016}, a quality which we and others leverage by casting the surrogate modeling problem as an image-to-image regression problem.

The challenge, however, is that such NN-based surrogate models must be trained on a sufficiently large dataset to enable generalization to unseen inputs. 
The value of these NN-based surrogate models thus hinges on the assumption that the cost of generating the training dataset does not exceed the cost of standard ensemble-based UQ approaches such as Monte Carlo analysis.\footnote{We mention Monte Carlo analysis throughout this manuscript. Unless otherwise specified, this should be interpreted as shorthand for Monte Carlo analysis using a traditional numerical solver.} As it turns out, this is a hefty assumption that is not supported by the field's current theoretical understanding of NN models.

To illustrate this, we will consider a typical setting for UQ. Suppose, as is common in many scientific applications, that we wish to analyze some output variable of a PDE on domain $\Omega$, whose true value is denoted by $Q:\Omega\rightarrow\mathbb{R}$, subject to a random input variable $\xi$ such that $Q=Q(\xi)$.\footnote{This extends to sets of random variables as well as random functions on the domain.} In particular, we aim to estimate $\mathbb{E}_{\xi}[Q]$.\footnote{Our analysis extends readily to other moments, and even full PDFs and CDFs.} Let $Q_M=Q_M(X_M)\in\mathbb{R}$ denote the approximation of $Q$ from the numerical solution to the PDE\footnote{For example, via finite difference, finite element, or finite volume methods} on a discretized computational domain of $M$ nodes, where $X_M:=(\xi_1,...,\xi_M)$ is the vector of random inputs on these nodes. We assume that $\mathbb{E}[Q_M]\rightarrow\mathbb{E}[Q]$ as $M\rightarrow\infty$.

Given a fixed data generation budget (for example, sufficient to run our PDE solver $N$ times, on $N$ samples of $X_M$), two possible approaches to estimating $\mathbb{E}_{\xi}[Q]$ are to construct a typical Monte Carlo estimator $\hat{Q}_{MC}$ with the $N$ solutions, or to train a NN-based surrogate model on the $N$ solutions and use the surrogate to estimate $\hat{Q}_{NN}$.
We denote our chosen estimator of $\mathbb{E}[Q_M]$ by $\hat{Q}_M$ and quantify the accuracy of this estimator via its mean square error (MSE) (also called the average discrepancy):
\begin{align}
    \mathcal D(\mathbb{E}[Q],\hat{Q}_M) &= \mathbb{E}[(\mathbb{E}[Q]-\hat{Q}_M)^2],
\end{align}
which can be decomposed into its bias-variance form as follows:
\begin{align*}
    \mathcal D(\mathbb{E}[Q], \hat{Q}_M) &=\mathbb{E}[(\mathbb{E}[Q]-\hat{Q}_M)^2]\\
    &=\mathbb{E}[Q]^2-2\mathbb{E}[\hat{Q}_M]\mathbb{E}[Q]+\mathbb{E}[\hat{Q}_M]^2-\mathbb{E}[\hat{Q}_M]^2+\mathbb{E}[\hat{Q}_M^2]\\
    &=\underbrace{\big(\mathbb{E}[\hat{Q}_M]-\mathbb{E}[Q]\big)^2}_{\text{bias}^2}+\underbrace{\mathbb{E}\big[(\hat{Q}_M-\mathbb{E}[\hat{Q}_M])^2\big]}_{\text{variance} = \epsilon^2_\text{samp}}.\\
\end{align*}
Since our estimator $\hat{Q}_M$ is based on a numerical approximation to $Q$, we can further decompose the bias term:
\begin{align*}
    \big(\mathbb{E}[\hat{Q}_M]-\mathbb{E}[Q]\big)^2 &=\Big(\big(\mathbb{E}[Q_M]-\mathbb{E}[Q]\big)+\big(\mathbb{E}[\hat{Q}_M]-\mathbb{E}[Q_M]\big)\Big)^2\\
    &=\underbrace{\big(\mathbb{E}[Q_M]-\mathbb{E}[Q]\big)^2}_{\epsilon^2_{\text{disc}}}+\underbrace{\big(\mathbb{E}[\hat{Q}_M]-\mathbb{E}[Q_M]\big)^2}_{\epsilon^2_{\text{est}}}-\underbrace{2\big(\mathbb{E}[\hat{Q}_M]-\mathbb{E}[Q_M]\big)\big(\mathbb{E}[Q_M]-\mathbb{E}[Q]\big)}_{\epsilon^2_*},
\end{align*}
where $\epsilon^2_{\text{disc}}$ is the bias due to discretization, $\epsilon^2_{\text{est}}$ is the bias of the estimator, and $\epsilon^2_*$ is a covariance term. Since MC estimators are inherently unbiased, for $\hat{Q}_M=\hat{Q}_{MC}$ we have $\mathbb{E}[\hat{Q}_{M}]=\mathbb{E}[Q_M]$ and thus:
\begin{align*}
    \mathcal D(\mathbb{E}[Q],\hat{Q}_{MC})&=\underbrace{\big(\mathbb{E}[Q_M]-\mathbb{E}[Q]\big)^2}_{\epsilon^2_{\text{disc}}}+\underbrace{\mathbb{E}\big[(\hat{Q}_{MC}-\mathbb{E}[\hat{Q}_{MC}])^2\big]}_{\epsilon^2_\text{samp}}.
\end{align*}
Therefore, for a suitably refined discretization, the average discrepany of an MC estimator for $\mathbb{E}[Q]$ decays inversely with the number of samples --- a well-known result. In contrast, our NN-based surrogate model estimator $\hat{Q}_{NN}$ cannot be assumed to be unbiased. We therefore have:
\begin{align*}
    \mathcal D(\mathbb{E}[Q], \hat{Q}_{NN})&=\underbrace{\big(\mathbb{E}[Q_M]-\mathbb{E}[Q]\big)^2}_{\epsilon^2_{\text{disc}}}+\underbrace{\big(\mathbb{E}[\hat{Q}_{NN}]-\mathbb{E}[Q_M]\big)^2}_{\epsilon^2_{\text{est}}}\\
    &-\underbrace{2\big(\mathbb{E}[\hat{Q}_{NN}]-\mathbb{E}[Q_M]\big)\big(\mathbb{E}[Q_M]-\mathbb{E}[Q]\big)}_{\epsilon^2_*}+\underbrace{\mathbb{E}\big[(\hat{Q}_{NN}-\mathbb{E}[\hat{Q}_{NN}])^2\big]}_{\epsilon^2_\text{samp}}
\end{align*}
where $\epsilon^2_*$ cannot be further specified without assumptions on the bias of the NN estimator and the correlation between the discretization error and estimator error.

Now let $\epsilon^2_{NN}=\epsilon^2_{\text{est}}+\epsilon^2_*+\epsilon^2_{\text{samp}}$ represent the overall error of the NN-based estimator.
The field's current understanding of neural scaling laws and recent empirical studies  \cite{baidu_scaling,kaplan_scaling_2020} suggest that the convergence of $\epsilon^2_{\text{NN}}$ with N is of $\mathcal{O}(N^{\alpha})$ with $\alpha\in\{-0.35, -0.07\}$, significantly slower than the $\mathcal{O}(-\frac{1}{2})$ rate of $\epsilon^2_{\text{samp}}$ for $\hat{Q}_{MC}$. Even in the specific sub-field of ``ML for science,'' both theoretical and empirical findings tend to yield a rate close to $\mathcal{O}(-\frac{1}{2})$ \cite{stuart_cost_accuracy, Convergence_theory_linear}, and in many cases much worse \cite{adcock2024optimaldeeplearningholomorphic}.
We therefore cannot expect to generate a generalizable machine-learned surrogate model using fewer training samples than would be required by traditional Monte Carlo analysis.


While NN-based surrogate models offer significant advantages over traditional MC analysis, this slow rate of convergence with respect to training set size presents a substantial barrier to their widespread adoption in the scientific community. We therefore argue that a necessary condition for the viability of NN-based surrogate models for the purpose of UQ is a dramatic reduction in the marginal cost of generating additional training samples. One strategy to achieve this is multifidelity learning, which aims to exploit the tradeoff between accuracy and cost (or abundance) of training data. Prior studies seeking to use multifidelity data to address the challenge of training data generation have primarily focused on generating a portion of the data on a coarser mesh \cite{song_transfer_2022, howard_multifidelity_2022} or with a less exact model \cite{howard_multifidelity_2022, meng_composite_2020}.

This work introduces a novel approach, whereby the lower-fidelity dataset is generated at a lower dimension rather than on a coarser mesh or with simpler physics. We thus build on the literature for transfer learning with multifidelity data while drawing inspiration from the curse of dimensionality. In the context of numerical solutions to PDEs, the curse of dimensionality refers to the exponential increase in the computational cost of a single solve for every increase in the PDE's dimension \cite{hutzenthaler_overcoming_2022}. We exploit the converse of this phenomenon to harness dramatic computational savings by generating a portion of the training data in a lower dimension. Then, for a fixed data generation budget, we are able to obtain far more training samples than would otherwise be possible. In a similar manner to multilevel Monte Carlo, this additional data serves to decrease $\epsilon^2_{NN}$ without affecting  $\epsilon^2_{\text{disc}}$.

In particular, we demonstrate our proposed approach by constructing a CNN-based PDE surrogate model for a 2D multiphase flow problem using a mixture of solutions to the 2D problem itself and solutions to its 1D approximation as training data. We achieve this using a dense encoder-decoder CNN architecture and a transfer learning strategy to incorporate the different classes of training data. We find that a CNN surrogate trained on our multi-dimensional dataset achieves lower error than a CNN surrogate trained on a 2D dataset with several times the data generation budget.  We then demonstrate the value of our surrogate model in a sample UQ task, where it outperforms the standard Monte Carlo approach despite its remarkably abbreviated training data budget.

\section{Methods}
We propose a novel training strategy for NN-based surrogate models of PDEs that dramatically reduces the cost of data generation. Using transfer learning, we demonstrate that a CNN surrogate can be trained on a mixture of numerical solutions to the original $d$-dimensional problem and its $(d-1)$-dimensional approximation. Thanks to the curse of dimensionality, this strategy yields dramatic computational savings that render NN-based surrogate modeling viable even in the face of poor convergence rates.

Our approach is an example of multifidelity machine learning, which leverages data spanning multiple levels of deviation from the true system of interest. While high-fidelity data is more accurate, low-fidelity data is more abundant or less computationally expensive to generate. Multifidelity learning, by utilizing both types of data, aims to strike a more favorable balance between accuracy and generalizability than is possible with either type of data alone for a fixed data generation budget. In particular, this is how our method is able to achieve such noteworthy efficiency gains; although a large number of training samples may be needed for generalization to unseen inputs, a large proportion of these samples can be generated very quickly.

A fundamental assumption underlying multifidelity learning is that the low-fidelity data, while less accurate than the high-fidelity data, still contains valuable information that can help the model learn the desired input-output mapping. A similar intuition underlies transfer learning, which operates on the premise that the learning performed for one modeling task is still valuable when applied to a new task. 
Alternative approaches to multifidelity learning, for example explicitly estimating the correlation between high- and low-fidelity data and embedding this into the model \cite{chen_feature-adjacent_2024}, have also proven successful. In the present work, the natural alignment between multifidelity learning and transfer learning motivates our use of transfer learning to implement our multifidelity approach.

In the remainder of this section, we briefly describe our model architecture and 
transfer learning strategy, as well as the computational test problem, data generation process, and computational resources we used to generate our results.

\subsection*{Model architecture} We take a similar approach to \cite{song_transfer_2022}, adapting the DenseED architecture proposed by \cite{zhu_bayesian_2018} to accommodate a transfer learning strategy. DenseED is a dense fully convolutional encoder-decoder CNN that has been successfully used to develop surrogate models for UQ in groundwater \cite{mo_deep_2019-1_deep} and multiphase flow \cite{mo_deep_2019_dcedn} problems.  

The architecture consists of two main components: the ``encoder'' and the ``decoder,'' each of which is composed of a series of alternating dense blocks and transition layers. The encoder's purpose is to extract and compress coarse features from the high-dimensional input, while the decoder acts to reconstruct the coarsened features into the output fields. The encoder-decoder architecture is a popular choice for image-to-image regression problems, as it excels in capturing complex spatial hierarchies and relationships within images.

The alternating dense blocks and transition layers work together to efficiently propagate information through the network while managing its complexity. The dense blocks contain skip connections between all layers, enhancing information flow and thereby reducing the amount of training data needed.\footnote{That is, relative to DNNs lacking connections between nonadjacent layers} Transition layers between the dense blocks perform convolutions, reduce the number of feature maps created by the dense blocks, and upsample or downsample depending on whether they lie in the encoder or decoder. 

\subsection*{Transfer learning} Transfer learning is a well-explored strategy \cite{transfer_shao,transfer_weiss_survey_2016} based on the premise that the knowledge gained from training one model can be applied when training a new model, even if for a completely different task. This can translate to directly applying a pre-trained model to a new prediction task, or, more commonly, using the weights from pre-trained model layers to initialize those of a new model. Our approach falls into the second category.

Specifically, we first train a low-fidelity model using the $(d-1)$-dimensional data, then use the resulting model to initialize a high-fidelity model trained on the $d$-dimensional data. This provides a straightforward means of incorporating both sets of training data, such that the bulk of the training is performed on the more abundant low-fidelity data, while the high-fidelity data is used to refine the model predictions. 

\begin{figure}
\centering
\includegraphics[width=\linewidth]{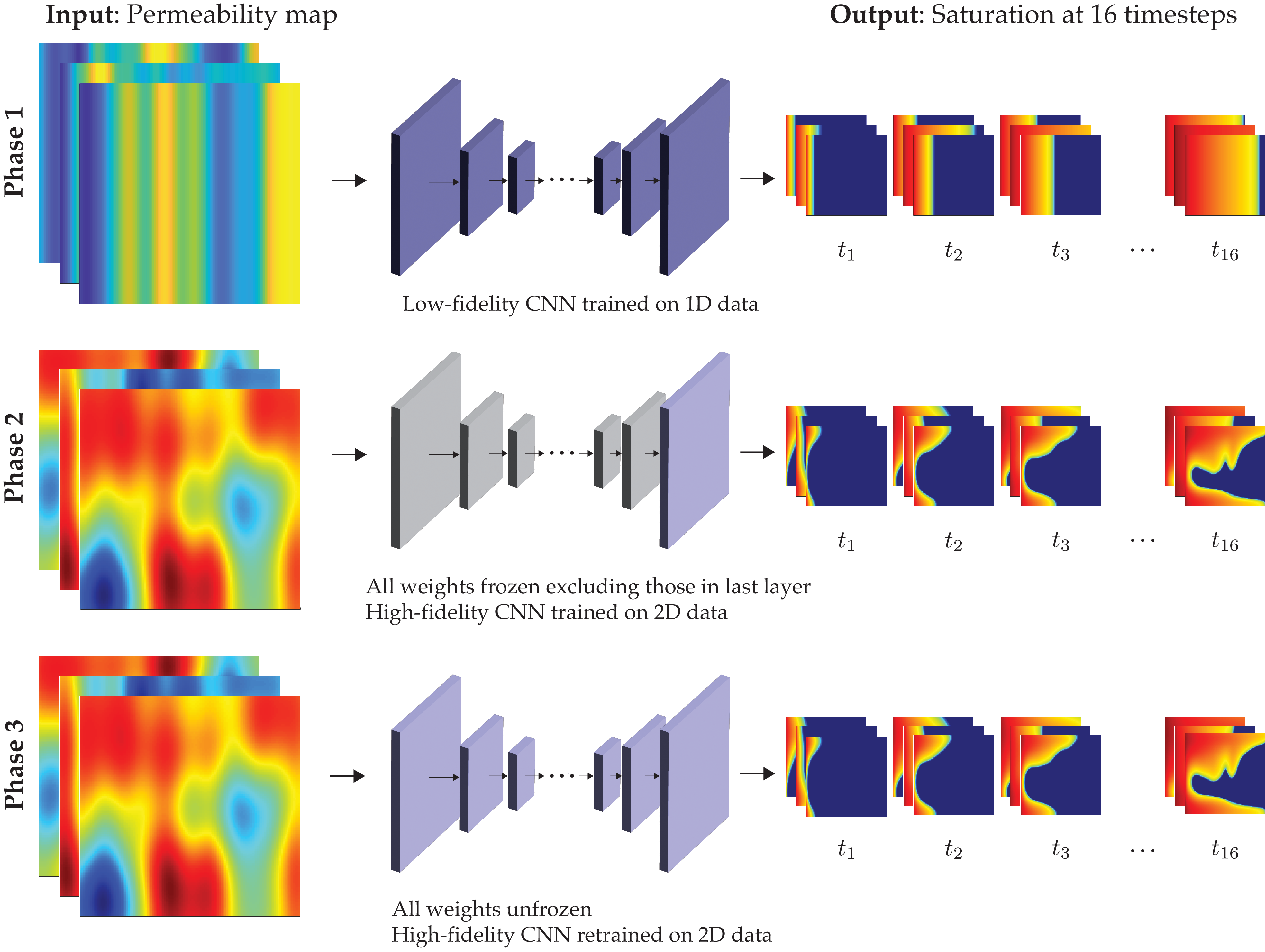}
\caption{We train our CNN surrogate in 3 phases. Phase 1 produces a low-fidelity surrogate trained exclusively on the 1D data, scaled up to 2D for compatibility. Phase 2 produces a high-fidelity surrogate in which all weights excluding those in the last layer are identical to those of the low-fidelity model, but the weights in the last layer are retrained with the 2D data. Phase 3 is initialized with the model weights from Phase 2, then all weights are unfrozen and retrained again with the 2D data, producing a refined high-fidelity surrogate model.}
\label{fig:phases}
\end{figure}

We implement this transfer learning strategy by training in three phases, building on the work of \cite{song_transfer_2022}. In Phase 1, we train a low-fidelity CNN surrogate exclusively on $(d-1)$-dimensional data. In Phase 2, we freeze all CNN weights excluding those in the final convolutional layer, which we retrain using the $d$-dimensional data. Finally, in Phase 3, we unfreeze all weights and retrain the entire network with the $d$-dimensional data. By initializing the weights in Phases 2 and 3 with those from the previous phase, our training strategy enables the model to leverage earlier training and reduces the number of high-fidelity $d$-dimensional training samples ultimately required. . Figure \ref{fig:phases} illustrates the three training phases, highlighting the respective inputs and outputs for the model at each stage.

\subsection*{Computational test problem}
To demonstrate our proposed method, we consider a multiphase Darcy flow problem for two incompressible and immiscible fluids, denoted by $l\in\{1,2\}$. This problem is pervasive in science and engineering applications and is particularly relevant for applications such as carbon sequestration.
The governing equations are the coupled PDEs relating saturation $S_l(\mathbf{x},t)$, velocity $\mathbf{v}_l(\mathbf{x},t)$, and pressure $P_l(\mathbf{x},t)$:
\begin{align}\label{eq:PDE}
&\phi\frac{\partial S_l}{\partial t} + \nabla\cdot\mathbf{v}_l+q_l=0,\quad\quad\mathbf{x}=(x_1,x_2)\in D, \quad t\in[0,T],\\
&\mathbf{v}_l=-k\frac{k_{rl}}{\mu_l}\nabla P_l.
\end{align}
Here, $\mu_l$ denotes viscosity, $q_l$ denotes forcing, and $k(\mathbf{x})$ denotes the uncertain intrinsic permeability. Apart from $k(\mathbf{x})$, all other model inputs are known and fixed. We assume porosity $\phi=0.25$, $P_1=P_2\equiv P(\mathbf{x},t)$, and adopt the Brooks-Corey constitutive model for relative permeability of the $l$th phase, $k_{rl}$, such that $k_{rl}=k_{rl}(S_l)$ \cite{brooks} (see Appendix \ref{sec:brooks} for additional details). The unknown outputs of interest are the saturation of each phase, $S_l(\mathbf{x},t)$. 

We consider a 2D, square domain $\Omega$ of size $150\times 150$ with Dirichlet boundary conditions on the left and right boundaries and no-flow conditions on the upper and lower boundaries:
\begin{align}
S_1 = 1.0,\quad&\mathbf{x}\in\Gamma_l,\\
P = 10.2,\quad&\mathbf{x}\in\Gamma_l,\\
P = 10.1,\quad&\mathbf{x}\in\Gamma_r,\\
\frac{\partial P}{\partial x_2}=0,\quad&\mathbf{x}\in\Gamma_b\cup\Gamma_t,
\end{align}
where $\Gamma$ with subscripts $l, r, b, t$ denotes the left, right, bottom, and top boundaries of $\Omega$, respectively.
We also impose the following initial condition for all $\mathbf{x}\in \Omega$:
\begin{align}
S_1(\mathbf{x},0)&=0,\\
P(\mathbf{x},0)&=10.1.
\end{align}

We model $k(\mathbf{x})$ as a second-order stationary random field. The exact specification for $k(\mathbf{x})$ is provided in Appendix \ref{sec:k}.

\subsection*{Data generation}\label{sec:data_gen}
Our training data consists of numerical solutions to this PDE obtained via a finite volume solver in MATLAB on a grid of size $150\times 150$. The high-fidelity data consists of the output to our 2D test problem, while the low-fidelity data consists of the output to the problem's 1D approximation. Running the MATLAB-based simulation on an Apple M1 Max chip with 32 GB of RAM, each 2D run takes 311 seconds to complete and each 1D run takes just over 1 second.

The 1D approximation of the computational test problem is computed by setting $\mathbf{x}\equiv(x_1)\in \Omega$ in equation \ref{eq:PDE}, where $\Omega$ is now a 1D pipe of length 150. We maintain the same initial conditions and the same boundary conditions on $\Gamma_l$ and $\Gamma_r$. We also regenerate the random input field $k(\mathbf{x})$ with the same statistical properties but in 1D.\footnote{This is a design choice. One could also elect to take ``slices'' of the random input field, but this is not guaranteed to maintain the statistical moments of the $d$-dimensional field.} For physics-based simulations such as the one explored here, the physics dictates which dimensions should be retained in the lower-dimensional approximation of the full system. In this case, the dynamics are dominated by horizontal flow from the left to the right, motivating the decision to model $x_1$ rather than $x_2$ in the 1D approximations. For problems in which the dynamics are not clearly dominated by one dimension over another, it may be beneficial to generate multiple sets of $(d-1)$-dimensional data, where each is generated by dropping one of the candidate dimensions.

\begin{figure}
\centering
\includegraphics[width=0.8\linewidth]{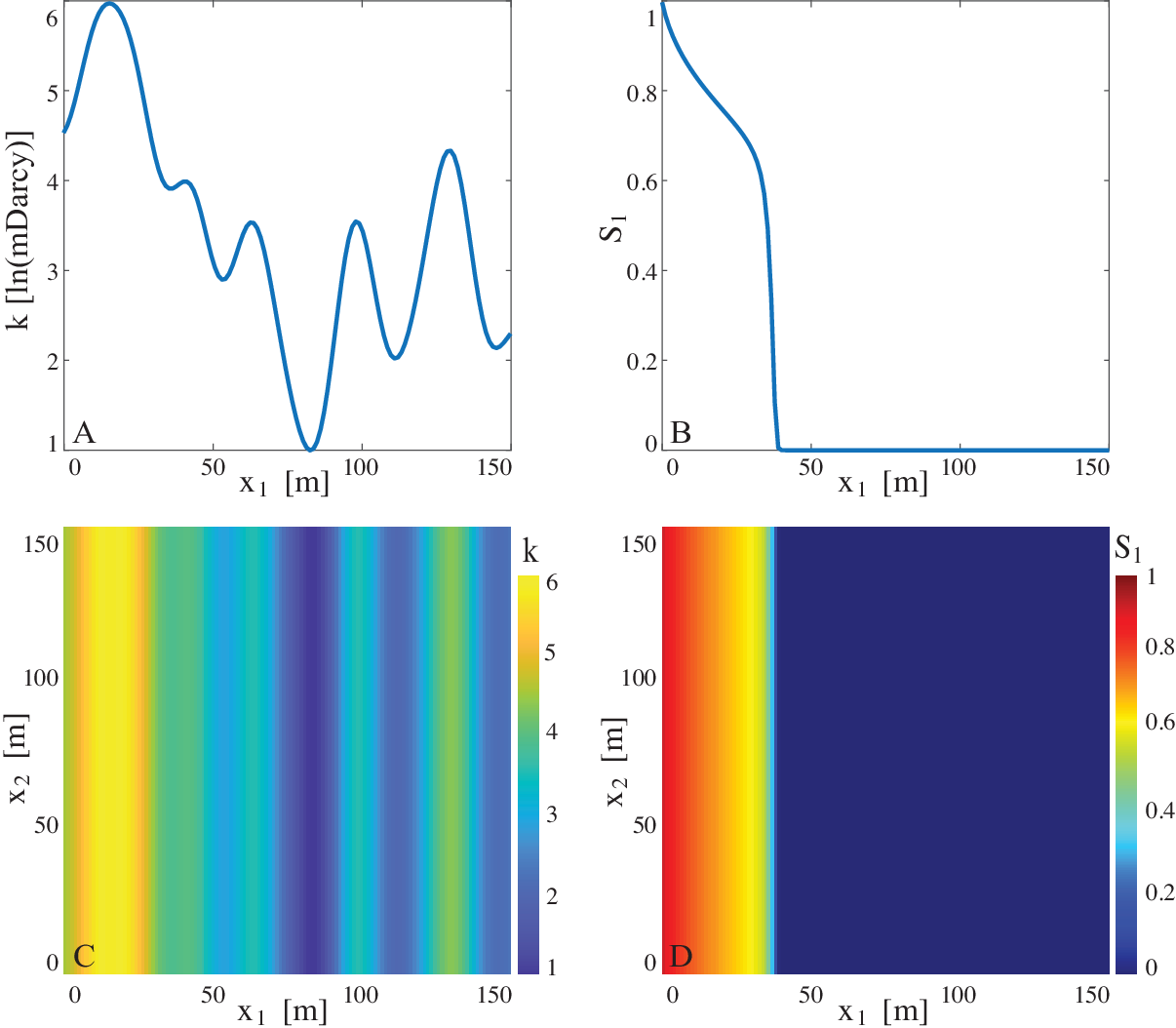}
\caption{For compatibility, we transform the 1D input-output pairs to 2D images. Here, we illustrate the ``low-frequency'' approach to 1D data generation, whereby each 1D input-output pair (panels A and B, respectively) is replicated $n=|x_2|$ times to generate a low-fidelity 2D input-output image pair (panels C and D, respectively). The y-axis values representing the magnitude of $k$ in panel A and $S_1$ in panel B are indicated by color in panels C and D.}
\label{fig:1Dto2D}
\end{figure}

Since our model architecture is designed to map 2D inputs to 2D outputs, we scale up the low-fidelity data (both inputs and outputs) to match the dimensions of the high-fidelity data --- in this case, $n\times n$.
We test two alternative methods to do this, which we refer to as the ``high-frequency'' and ``low-frequency'' approaches. In the high-frequency approach, we concatenate $n$ separate 1D permeability fields to generate a single input image and $n$ corresponding 1D solutions to generate the corresponding output image. In the low-frequency approach, we replicate a single 1D permeability field $n$ times to generate an input image, and replicate the corresponding 1D solution $n$ times to generate the output image. An example of an input-output image pair for the low frequency approach is provided in Figure \ref{fig:1Dto2D}. Note that this design choice affects the training dataset for Phase 1 but not Phase 2 or Phase 3.

\subsection*{Training and computational requirements}\label{sec:requirements}
We implemented our CNN surrogate using Python and PyTorch \cite{paszke2019pytorch} supporting CUDA. We conducted all training and experiments on the \texttt{gpu} partition of the Sherlock high-performance computing cluster. The GPU resources available in this partition include Tesla P40, P100, and V100 GPUs.

We used the popular Adam optimizer \cite{KingBa15} to minimize the $L_1$ loss and a separate test set of 100 samples or 25\% of the training set size, whichever was greater. Our hyperparameter values were informed by the work of \cite{song_transfer_2022} with no additional hyperparameter tuning. Specifically, we used an initial learning rate of 5e-4 in Phase 1, 5e-5 in Phase 2, and 1e-5 in Phase 3; 100 epochs in each phase; weight decay of 1e-5; factor of 0.6; minimum learning rate of 1.5e-06 in Phases 1 and 2 and 5e-07 in Phase 3. All other parameters were left to their default values.

\section{Results}
In this section we present three major insights gleaned from our numerical experiments. The first pertains to the viability of the method: we find that our surrogate model outperforms standard Monte Carlo in uncertainty quantification tasks, even with less data. We also present two findings relevant to the generation and setup of the training dataset.

\paragraph*{Our multifidelity CNN surrogate outperformed Monte Carlo in UQ.}
To evaluate the effectiveness of our model, we applied it to a sample uncertainty quantification task. A common challenge in hydrology and reservoir engineering is predicting the breakthrough time of an invading fluid, which is critical for estimating when pollutants might reach drinking water sources or when injected fluids will arrive at production wells \cite{breakthrough}.

\begin{figure}
\centering
\includegraphics[width=\linewidth]{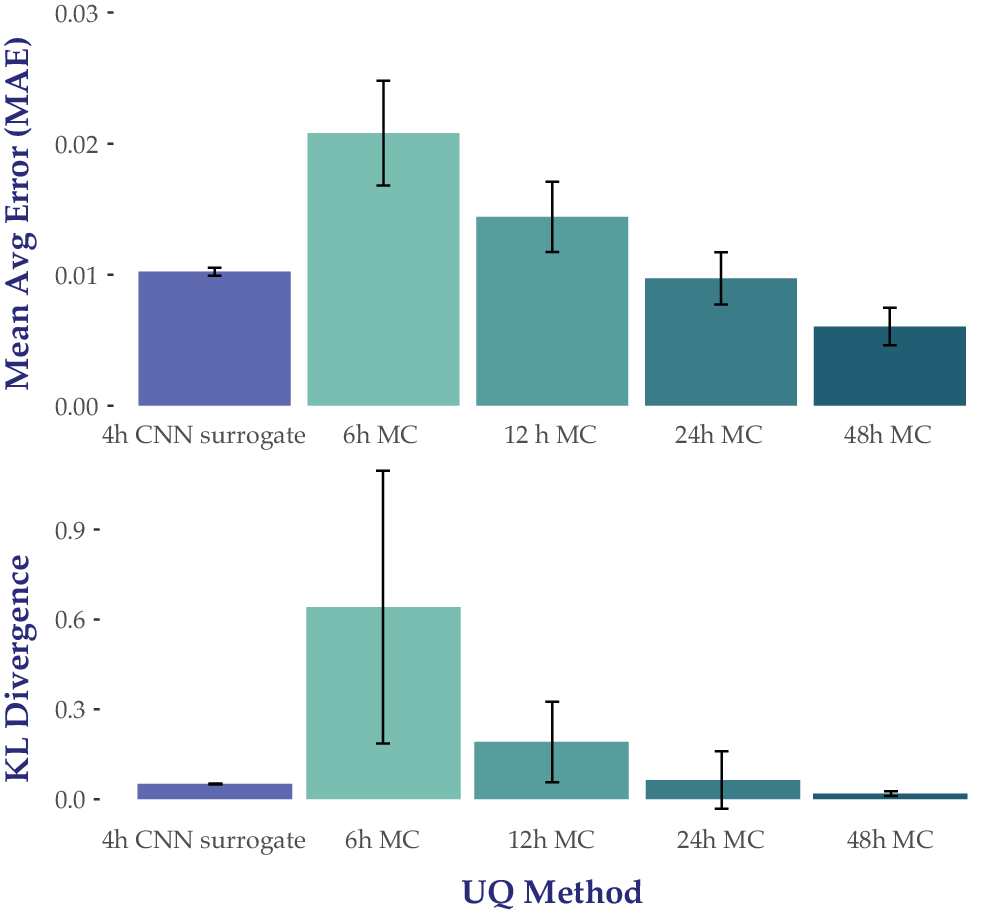}
\caption{UQ results: mean average error (MAE) and Kullback-Leibler (KL) divergence of the PDF of $T_b$. From left: 2000 forward passes of the CNN surrogate trained on 4 hours of multi-dimensional data (in purple); standard Monte Carlo with 6, 12, 24, and 48 hours of high fidelity runs (in progressively darker green). For both measures, the CNN surrogate achieves similar or better results than the Monte Carlo method on 24 hours of high fidelity runs, with lower variance. Error bars represent the standard deviation over 50 trials.}
\label{fig:UQ}
\end{figure}

We let $T_b$ denote the breakthrough time:
\begin{align*}
    T_b = \underset{t}{\arg\min} S_1(x_1=100,t)\geq 0.15,
\end{align*}
defined as the first time at which the invading phase reaches a desired saturation threshold (in this case, 0.15) at a given point in the domain (here, $x_1=100$). We approximated the PDF of $T_b$ by evaluating 2000 forward passes of our multifidelty CNN surrogate. This process took less than 18 seconds to complete once the model was trained. We compared the output to the PDF obtained from standard Monte Carlo with as many forward passes of the original high-fidelity numerical solver as possible for a given data generation budget. We found that the CNN surrogate, trained on only 4 hours' worth of multifidelity data, performed comparably to Monte Carlo on 24-48 hours' worth of high-fidelity numerical solutions. This result was robust to multiple error metrics, as depicted in Figure \ref{fig:UQ}.

\begin{figure}
\centering
\includegraphics[width=\linewidth]{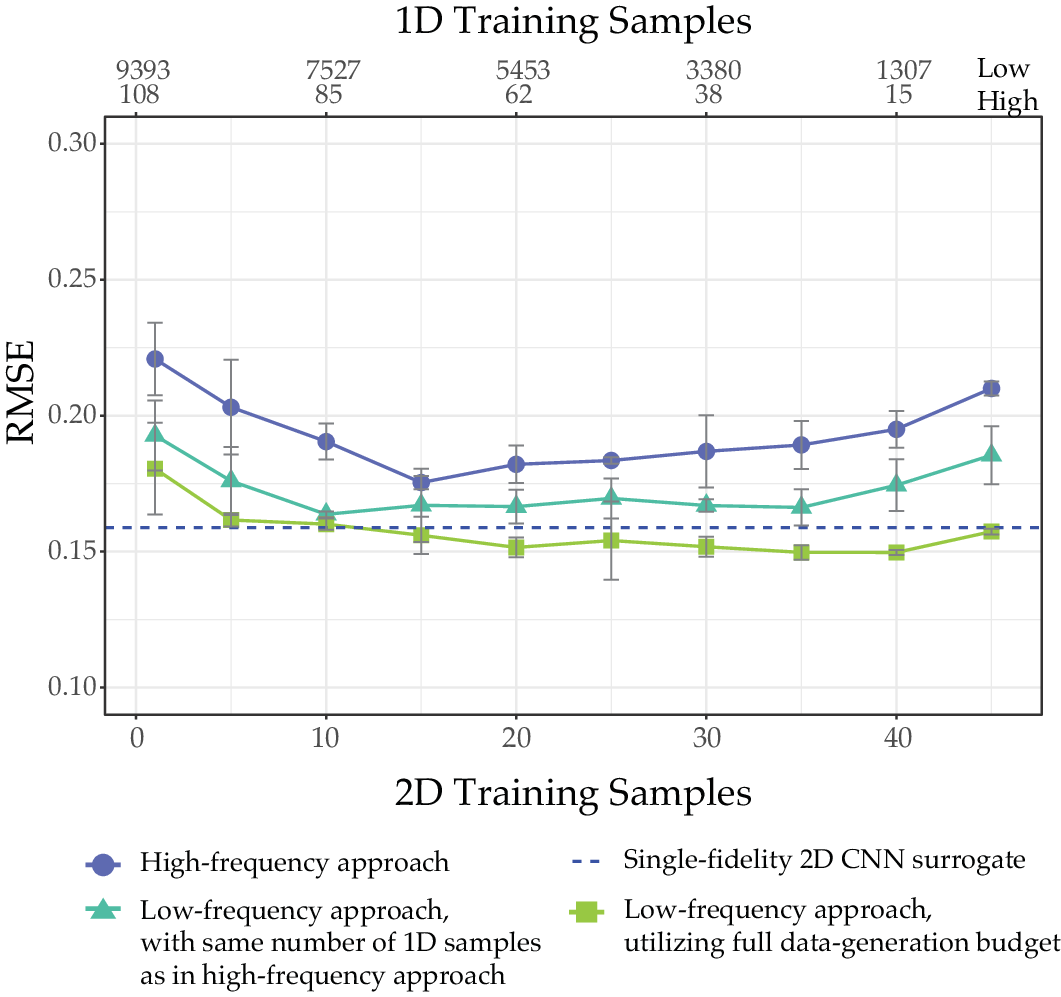}
\caption{RMSE for CNN surrogate models trained with different strategies, as a function of the number of PDE solutions in the training set (x-axis). The results here represent the average of the top 5 out of 30 runs for each training scenario, accompanied by error bars indicating two standard deviations in each direction. The dotted line indicates the error achieved by training a single-fidelity CNN surrogate on 12 hours' worth of 2D data, and the solid curves represent different training strategies for the multifidelity CNN surrogate. Each solid line represents results from training on 4 hours' worth of data across different allocations between 1D and 2D data. The purple curve was generated using high-frequency 1D training images. The blue curve was generated using low-frequency 1D training images, but the same total number of 1D images as in the high frequency approach (purple curve). This allows us to directly compare the relative benefit of the high- versus low-frequency approach to generating 1D training data. The green curve was generated using the maximum number of low-frequency 1D training images permitted by the data generation budget, for a fixed number of 2D images (so the number of 1D training images was approximately 88 times the amount used for the blue curve, since we conservatively assumed that each low-frequency sample took 1.5 seconds to generate, while each high-frequency sample took 133 seconds). Thus, the purple and blue curves were generated using different types of 1D images, but the same quantity. The blue and green curves were generated using the same type of 1D images, but different quantities. We clearly see that the low-frequency training data produces better results than the high-frequency training data, and that increasing the number of training samples improves performance significantly.
}
\label{fig:RMSE}
\end{figure}

\paragraph*{Balance between high- and low-fidelity data optimized performance.}
We found that balancing the training data generation budget between low- and high-fidelity data was optimal from the perspective of minimizing the root mean square error (RMSE). At the two extremes of: 1) nearly all high-fidelity data (and relatively few training samples); and 2) nearly all low-fidelity data (and relatively many training samples), the models performed worse than when the data generation budget was more evenly split. This result is illustrated in Figure \ref{fig:RMSE}, where the x-axis indicates an increasing number of 2D samples (bottom axis) and decreasing number of 1D samples (top axis) from left to right.  This result was also consistent across both the high- and low-frequency approaches to generating low-fidelity data. This finding is consistent with prior work \cite{song_transfer_2022}, and suggests that balancing 1D and 2D data in the training set allows the model to benefit from both the accuracy of the high-fidelity 2D data and the larger sample size afforded by the low-fidelity 1D data. 

Our multifidelity approach also addresses other drawbacks to exploiting an increased training set size. In particular, training models on larger datasets requires more time and storage capacity. Given these factors and the results presented above, the division of the data generation budget between high- and low-fidelity data should be determined depending on the user's needs and resources.

\paragraph*{The low-frequency approach to data generation performed significantly better than the high-frequency approach.}

Recall that our training data consist of input/output image pairs, the input being a permeability map, and the output being the set of saturation maps at each timestep. For compatibility, we scale up the 1D permeability and saturation maps to form 2D images. The method by which the 1D input/output pairs are scaled up to 2D images is a design choice. We tested two approaches, as discussed in Section \ref{sec:data_gen}. The first is to generate a 2D input/output image pair from $n$ 1D input/output pairs (where our domain is of size $n\times n$). The second is to generate a 2D input/output image pair from a single 1D input/output pair.  Approach 1 is achieved by concatenating $n$ distinct 1D permeability maps and $n$ corresponding outputs. Approach 2 (illustrated in Figure \ref{fig:1Dto2D}) is accomplished by replicating a single input and output $n$ times, where $n$ is the size of the domain in the second dimension. We refer to Approaches 1 and 2 as the high- and low-frequency approaches, respectively, corresponding to the frequency of the content they each contain. The high-frequency images take $n$ times as long to generate as the low-frequency images, but, in theory, comprise $n$ independent wavefronts and thus $n$ times as much information.

We found that the low-frequency approach to data generation significantly outperformed the high-frequency approach, despite containing less information and taking a fraction of the time to generate (just over 1 second versus 133 seconds). This can be seen in Figure \ref{fig:RMSE}, where the purple and blue curves show the high- and low-frequency results, respectively, each trained with the same number of high- and low-fidelity training samples.\footnote{For the low-frequency results, this means that data-generation did not utilize the full allotted budget of 4 hours.} This behavior is consistent with the well-known spectral bias of neural networks \cite{rahaman_spectral_2019, cai_multi-scale_2019}. These results are also consistent with our expectations --- given that our ultimate goal is to model a single flow scenario, 
the weights learned from modeling multiple flow scenarios may not be as relevant as those learned from modeling a single flow scenario.

We also found that increasing the number of low-frequency low-fidelity training samples to fully utilize the 4 hour data-generation budget further improved the performance of the surrogate. In fact, for several different splits between 1D and 2D data, our multifidelity model achieved lower error than even the CNN surrogate trained on 12 hours' worth of 2D data. These findings are shown by the green curve in Figure~\ref{fig:RMSE}. These results are encouraging as they indicate that supplementing a limited data generation budget with inexpensive ($d-1$)-dimensional data can lead to substantial improvements in model performance.

\section{Discussion}
In this work, we demonstrate a novel approach to training CNN-based surrogate models that addresses a fundamental but often-ignored challenge with such methods: slow convergence with respect to training dataset size. Specifically, we demonstrate that a CNN surrogate model for a 2D PDE can be trained on a mixture of numerical solutions to the 2D PDE and numerical solutions to its 1D approximation. With this approach, we achieve lower error in UQ tasks compared to Monte Carlo using several times the data generation budget. These results suggest that the curse of dimensionality may be tamed and even exploited in our favor; the ability of CNNs to learn useful features from a lower-dimensional approximation to the original problem allows us to generate much more training data than would otherwise be possible, given a fixed data generation budget and the typically high cost of running physics-based simulations.

We note that the results described here are achieved with a relatively off-the-shelf method. We selected the current model architecture based on its success in generating a surrogate model for similar problems \cite{song_transfer_2022, zhu_bayesian_2018, mo_deep_2019-1_deep, mo_deep_2019_dcedn}, but acknowledge that several measures could be taken to improve its performance. For example, we did not conduct any hyper-parameter optimization, and rather adopted the parameters identified by \cite{song_transfer_2022}. Data augmentation (e.g., adding rotated or flipped versions of training images to the training dataset as if they were independent samples) has been shown to improve NN performance \cite{gu_recent_2018}. This technique could potentially improve the ability of the ($d-1$)-dimensional dataset to capture dynamics in the missing dimension, or augment a small $d$-dimensional dataset. In terms of architecture, other studies on multifidelity learning emphasize the benefits of computing and exploiting the correlation between high- and low-fidelity data \cite{howard_multifidelity_2022, meng_composite_2020,  chen_feature-adjacent_2024}. Implementing memory-aware synapses, as in \cite{howard_multifidelity_2023}, could also ensure that the weights of the optimal layer are being retrained in each phase, so that useful weights are maintained and less useful weights are refined. Finally, we noted the occurrence of aliasing in the output images. While this is not a major issue for tasks like uncertainty quantification, techniques such as Fourier feature embedding \cite{NEURIPS2020_55053683_fourier} can help address this phenomenon, further improving performance and accuracy in capturing the sharp saturation front.

Our results suggest that it may be possible to tame the curse of dimensionality by training a CNN-based surrogate model on multi-dimensional data.
Our approach harnesses the strength of CNNs in capturing spatial relationships across multiple dimensions, while overcoming the slow convergence of NN-based surrogates with respect to training set size --- one of the fundamental challenges with NN-based prediction. While further study is needed to verify this approach on higher-dimensional problems and additional classes of PDEs, we expect that our results should readily extend to these regimes. The continued development of simple-yet-effective approaches such as the one proposed here could significantly broaden the applicability of CNN-based surrogate models, providing powerful tools for researchers dealing with high-dimensional UQ tasks across various scientific and engineering domains.

\section{Acknowledgements}
This work is supported by the U.S. Department of Energy, Advanced Scientific Computing Research program, under the Scalable, Efficient and Accelerated Causal Reasoning Operators, Graphs and Spikes for Earth and Embedded Systems (SEA-CROGS) project (Project No. 80278) and by the Stanford Graduate Fellowship (SGF) and Enhancing Diversity in Graduate Education (EDGE) Doctoral Fellowship Program. 

The authors gratefully acknowledge Dong Song, Sebastian Bosma, Amanda Howard, and Panos Stinis for their insightful comments and useful discussions.

\bibliographystyle{unsrtnat}
\bibliography{references_TL_MFD}

\appendix

\section{Appendix / supplemental material}

\subsection{Specification of random permeability field $k(\mathbf{x})$}\label{sec:k}

We model the uncertain intrinsic permeability on the domain, $k(\mathbf{x})$, using a second-order stationary random field. We let $$Y(\mathbf{x})=\ln k$$
be multivariate Gaussian with mean $\langle Y\rangle=0$, variance $\sigma^2_Y=2.0$, correlation length $\lambda_Y=19,$ and exponential two-point covariance $C(\mathbf{x},\mathbf{y})=\sigma^2_Y\exp(-|\mathbf{x}-\mathbf{y}|/\lambda_Y)$. Following the work of \cite{taverniers}, we use a truncated Karhunen-Lo\'{e}ve expansion with $p=31$ terms to represent $Y(\mathbf{x})$.

\subsection{Brooks-Corey constitutive model for relative permeability}\label{sec:brooks}

We adopt the Brooks-Corey constitutive model for relative permeability of the $l$th phase, $k_{rl}$, such that $k_{rl}=k_{rl}(S_l)$ \citep{brooks}. This model is widely accepted in the hydrology and reservoir engineering community.

The Brooks-Corey model computes $k_{rl}=k_{rl}(S_l)$ using the following relation:
\begin{align}
k_{rl}(S_l) = \frac{S_l-S_l^{\text{r}}}{1-\sum_l S_l^{\text{r}}},
\end{align}
where $S_l$ is the saturation of the $l$th fluid phase and $S_l^{\text{r}}$ is the residual saturation of the $l$th fluid phase.

\end{document}